\pdfoutput=1
\documentclass[11pt,a4paper]{article}
\usepackage{booktabs} 
\usepackage{multirow}

\usepackage{amsmath,amsfonts,bm}

\def\eqref#1{equation~\ref{#1}}

\def\1{\bm{1}}

\DeclareMathAlphabet{\mathsfit}{\encodingdefault}{\sfdefault}{m}{sl}
\SetMathAlphabet{\mathsfit}{bold}{\encodingdefault}{\sfdefault}{bx}{n}

\DeclareMathOperator*{\argmin}{arg\,min}

\usepackage[hyperref]{emnlp2020}
\usepackage{times}
\usepackage{latexsym}

\usepackage{adjustbox}

\usepackage{subfig}
\usepackage{microtype}

\usepackage{multicol}

\usepackage{graphicx}
\usepackage[normalem]{ulem}
\usepackage{amsmath} 
\usepackage{booktabs} 

\newcommand{\T}{{\scriptscriptstyle \top}}

\aclfinalcopy 
\def\aclpaperid{0020} 

\title{Unsupervised Distillation of Syntactic Information \\from Contextualized Word Representations}

\author{Shauli Ravfogel$\thanks{~~Equal contribution}\,\,\,$\textsuperscript{1,3} \, Yanai Elazar$\footnotemark[1]\,\,\,$\textsuperscript{1,3} \, Jacob Goldberger \textsuperscript{2} \, Yoav Goldberg\textsuperscript{1,3} \\
\textsuperscript{1}Computer Science Department, Bar Ilan University \\
\textsuperscript{2}Faculty of Engineering, Bar Ilan University \\
\textsuperscript{3}Allen Institute for Artificial Intelligence \\
  {\tt  \{shauli.ravfogel, yanaiela,yoav.goldberg\}@gmail.com}\\
  \tt jacob.goldberger@biu.ac.il
  }

\date{}

\begin{document}
\maketitle
\begin{abstract}
Contextualized word representations, such as ELMo and BERT, were shown to perform well on various semantic and syntactic tasks. In this work, we tackle the task of unsupervised disentanglement between semantics and structure in neural language representations: we aim to learn a transformation of the contextualized vectors, that discards the lexical semantics, but keeps the structural information. To this end, we automatically generate groups of sentences which are structurally similar but semantically different, and use metric-learning approach to learn a transformation that emphasizes the structural component that is encoded in the vectors. We demonstrate that our transformation clusters vectors in space by structural properties, rather than by lexical semantics. Finally, we demonstrate the utility of our distilled representations by showing that they outperform the original contextualized representations in a few-shot parsing setting.
\end{abstract}

\section{Introduction}

Human language\footnote{In this work we focus on English.} is a complex system, involving an intricate interplay between meaning (“semantics”) and structural rules between words and phrases (“syntax”). Self-supervised neural sequence models for text trained with a language modeling objective, such as ELMo \citep{elmo}, BERT \citep{bert}, and RoBERTA \citep{roberta}, were shown to produce representations that excel in recovering both structure-related information \citep{gulordava2018LMagreement, vanschijndel2018gardenpath, wilcox2018fillergap, goldberg2019} as well as in semantic information \citep{bert-questions, joshi2019bert}. 

\begin{figure}[t]
    \centering
    \includegraphics[width=1\linewidth]{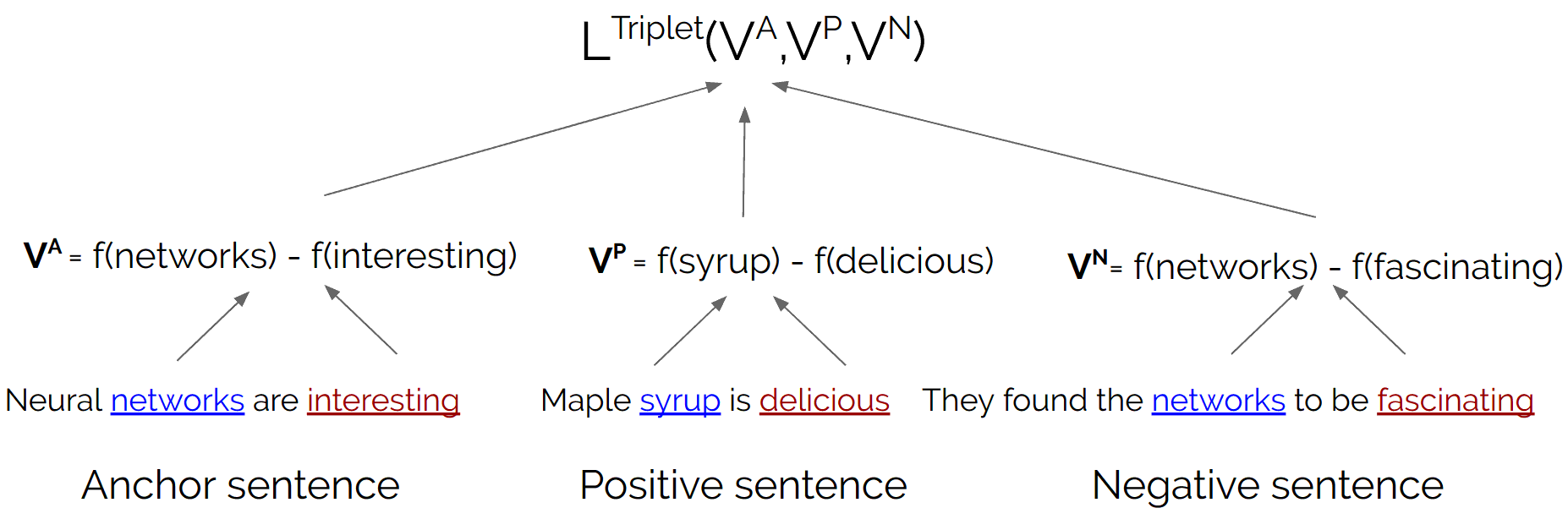}
    \caption{An illustration of triplet-loss calculation. Pairs of words are represented by the difference between their transformation $f$, which is identical for all words. The pairs of words in the anchor and positive sentences are lexically different, but structurally similar. The negative example presented here is especially challenging, as it is lexically similar, but structurally different.}
    \label{fig:triplet-loss}
\end{figure}

In this work, we study the problem of disentangling structure from semantics in
neural language representations: we aim to extract representations that capture
the structural function of words and sentences, but which are not sensitive to
their content. For example, consider the sentences:
    
\begin{enumerate}
    \item Neural networks are interesting.
    \item I study neural networks.
    \item Maple syrup is delicious.
    \item John loves maple syrup. 
\end{enumerate}

\noindent
While (1) and (3) are different in content, they share a similar structure, the corresponding words in them, while unrelated in meaning,\footnote{We focus on lexical semantics.} serve the same function. Similarly for sentences (2) and (4). 
In contrast, sentence (1) shares the phrase \emph{neural networks} with sentence (2), and \emph{maple syrup} is shared between (3) and (4).\footnote{There is a syntactic distinction between the two, with ``maple" being part of a noun compound and ``neural" being an adjective. However, we focus in their similarity as noun modifiers in both phrases.} While the two occurrences of each phrase share the meaning, they are used in different structural (syntactic) configurations, serving different roles within the sentence (appearing in subject vs object position).\footnote{These differences in syntactic position are also of relevance to language modeling, as different positions may pose different restrictions on the words that can appear in them.} 
We seek a representation that will expose the similarity between ``networks''
in (1) and ``syrup'' in (2), while ignoring the similarity between ``syrup''
in (2) and ``syrup'' in (4).

We seek a function from contextualized word representations to a space
that exposes these similarities.
Crucially, we aim to do this in an unsupervised manner: we do not want to inform the process of the kind of structural information we want to obtain. We do this by learning a transformation that attempts to remove the lexical-semantic information in a sentence, while trying to preserve structural properties.

 Disentangling syntax from lexical semantics in word representations is a desired property for several reasons. From a purely scientific perspective, once disentanglement is achieved, one can better control for confounding factors and analyze the knowledge the model acquires, e.g.  attributing the predictions of the model to one factor of variation while controlling for the other. In addition to explaining model predictions, such disentanglement can be useful for the comparison of the representations the model acquires to linguistic knowledge. From a more practical perspective, disentanglement can be a first step toward controlled generation/paraphrasing that considers only aspects of the \textit{structure}, akin to the style-transfer works in computer vision, i.e., rewriting a sentence while preserving its structural properties while ignoring its \textit{meaning}, or vice-versa. It can also inform search-based application in which one can search for ``similar'' texts while controlling various aspects of the desired similarity.

To achieve this goal, we begin with the intuition that the structural component in the representation (capturing the \emph{form}) should remain the same regardless of the lexical semantics of the sentence (the \emph{meaning}). Rather than beginning with a parsed corpus, we automatically generate a large number of structurally-similar sentences, without presupposing their formal structure (\S\ref{sec:sentences-generation}). This allows us to pose the disentanglement problem as a metric-learning problem: we aim to learn a transformation of the contextualized representation, which is \textit{invariant} to changes in the lexical semantics within each group of structurally-similar sentences (\S\ref{sec:triplet-loss}). We demonstrate the structural properties captured by the resulting representations in multiple experiments (\S\ref{sec:experiments}), among them automatic identification of structurally-similar words and few-shot parsing. 

 We release our code at \url{https://github.com/shauli-ravfogel/NeuralDecomposition}.

\section{Related Work}

The problem of disentangling different sources of variation has long been studied in computer vision, and was recently applied to neural models \citep{bengio-representation-learning, lecum-disentangling, hadad-disentanglement}. Such disentanglement can assist in learning representations that are invariant to specific factors, such as pose-invariant face-recognition \citep{pose-invariant} or style-invariant digit recognition \citep{digits-style}. From a generative point of view, disentanglement can be used to modify one aspect of the input (e.g., ``style''), while keeping the other factors (e.g., ``content'') intact, as done in neural image style-transfer \citep{image-style-transfer}. 

 
In NLP, disentanglement is much less researched. In controlled natural language generation and style transfer, several works attempted to disentangle factors of variation such as sentiment or age of the writer, with the intention to control for those factors and generate new sentences with specific properties \citep{sohn2015learning, ficler2017controlling, lample2018multiple}, or transfer existing sentences to similar sentences that differ only in the those properties. The latter goal of style transfer is often realized by learning representations which are invariant to the controlled attributes \citep{fu2018style, HuControlledGeneration}.




Another main line of work which is relevant to our approach is that of probing. The concept, originally introduced by  \citet{diagnostic1} and  \citet{diagnostic2}, relies on training classifiers (probes) to expose symbolic linguistic information that is encoded in the model. A large body of works have shown sensitivity to both semantic \cite{nlp-bert-pipeline-tenney,richardson2019probing} and syntactic \cite{tenney2018what,lin2019open,reif2019visualizing,structural-probe,liu2019linguistic} information. \citet{structural-probe} demonstrated that it is possible to train a linear transformation, under which squared euclidean distance between transformed contextualized word vectors correspond to the distances between the
respective words in the syntactic tree. \citet{Eisner-IB} have used a
variational estimation method \citep{variational-IB} of the
information-bottleneck principle \citep{tishby2000information} to extract word embeddings that are useful to the end task of parsing. 


While impressive, those works presuppose a specific syntactic structure (e.g. annotated parse tree) and use this linguistic signal to learn the probe in a supervised manner. This approach can introduce confounding between \emph{extracting} information and \emph{learning} it by the probe \citep{Hewitt-control, probing-probing, probe-parser,amnesic-probing:2020}. In contrast, we aim to \emph{expose} the structural
information encoded in the network in an unsupervised manner, without
pre-supposing an existing syntactic annotation scheme. 

\section{Method}


Our goal is to learn a function $f : \mathbb{R}^n \mapsto \mathbb{R}^m$, which operates on contextualized word representations $x$ and extracts vectors $f(x)$ which make the structural information encoded in $x$ more salient, while discarding as much lexical information as possible. In the sentences ``Maple syrup is delicious'' and ``Neural networks are interesting'',  we want to learn a function $f$ such that $f(v^3_{\text{syrup}})$ $\approx$ $f(v^1_{\text{networks}})$, where $v^i_{\text{word}}$ is the contextualized vector representation of the word in sentence $i$. We also want 
$f(v^4_{\text{syrup}})$ $\approx$ $f(v^2_{\text{networks}})$, while keeping
$f(v^1_{\text{networks}})$ $\not\approx$ $f(v^2_{\text{networks}})$.

Moreover, we would like the \textit{relation} between the words ``maple'' and ``delicious'' in the third sentence, to be similar to the relation between ``neural'' and ``interesting'' in the first sentence: $\text{pair}(v^3_\text{maple},v^3_\text{delicious}) \approx \text{pair}(v^1_\text{neural},v^1_\text{interesting})$. Operatively, we represent pairs of words $(x,y)$ by the difference between their transformation $f(x) - f(y)$, and aim to learn a function $f$ that preserves:
$f(v^3_\text{maple})-f(v^3_\text{delicious}) \approx f(v^1_\text{neural})-f(v^1_\text{interesting})$. The choice to represent pairs this way was inspired by several works that demonstrated that nontrivial semantic and syntactic relations between uncontextualized word representations can be approximated by simple vector arithmetic \citep{mikolov2013distributed, mikolov2013linguistic, levy2014linguistic}.

To learn $f$, we start with groups of sentences such that the sentences within each group are known to share structure but differ in lexical semantics. We call the sentences in each group \emph{structurally equivalent}. Figure \ref{fig:equivalent-sentences} shows an example of two structurally equivalent sets. Acquiring such sets is challenging, especially if we do not assume a known syntactic formalism and cannot mine for sentences based on their observed tree structures. To this end, we automatically generate the sets starting with known sentences and sampling variants from a language model (\S\ref{sec:sentences-generation}).  
Our sentence-set generation procedure ensures that words from the same set that share an index also share their structural function. We call such words \emph{corresponding}.


We now proceed to learn a function $f$ to map contextualized vectors of corresponding words (and the relations between them, as described above) to neighbouring points in the space.


We train $f$ such that the representation assigned to positive pairs --- pairs that share indices and come from the same equivalent set --- is distinguished from the representations of negative pairs --- challenging pairs that come from different sentences, and thus do not share the structure of the original pair, but can, potentially, share their lexical meaning. We do so using Triplet loss, which pushes the representations of pairs coming from the same group closer together (\S\ref{sec:triplet-loss}). Figure \ref{fig:triplet-loss} sketches the network.

\subsection{Generating Structurally-similar Sentences}
\label{sec:sentences-generation}

\begin{figure*}[t!]
\centering
\begin{multicols}{2}
    \includegraphics[width=1\textwidth]{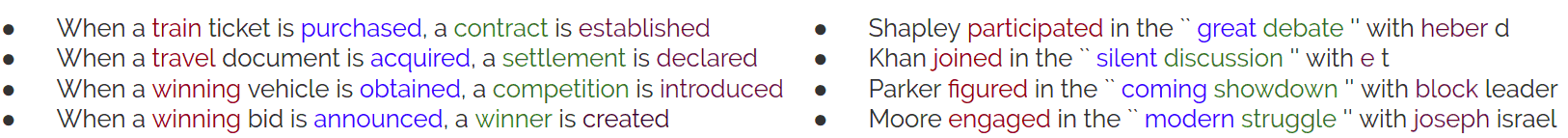}
    \end{multicols}
\caption{Two groups of structurally-equivalent sentences. In each group, the first sentence is original sentence from Wikipedia, and the sentences below it were generated by the process of repeated BERT substitution. Some sets of corresponding words--that is, words that share the same structural function--are highlighted in the same color.}
\label{fig:equivalent-sentences}
\end{figure*}

In order to generate sentences that approximately share their structure, we sequentially replace content words in the sentence with other content words, while aiming to maintain the grammatically of the sentence, and keep its structure intact. 

Since we do not want to rely on syntactic annotation 
when performing this replacement, we opted to use a pre-trained language model -- BERT -- under the assumption that strong neural language models do implicitly encode many of the syntactic restrictions that apply to words in different grammatical functions (e.g., we assume that BERT would not predict a transitive verb in the place of an intransitive verb, or a verb that accepts a complement in the place of a verb that does not accept a complement). While this assumption seems to hold with regard to basic distinctions such as transitive vs. intransitive verbs, its validity is less clear in the more nuanced cases, in which small differences in the surface level can translate to substantial differences in abstract syntactic structure -- such as replacing a control verb with a raising verb. This is a limitation of the current approach, although we find that the average sentence we generate is grammatical and similar in structure to the original sentence. Moreover, as our goal is to \emph{expose} the structural similarity encoded in neural language models, we find it reasonable to only capture the distinctions that are captured by modern language models.

\paragraph{Implementation} We start each group with a Wikipedia sentence, for which we generate $k=6$  equivalent sentences by iterating over the sentence from left to right sequentially, masking the ith word, and replacing it with one of BERT's top-30 predictions. To increase semantic variability, we perform the replacement in place (online): after randomly choosing a guess $w$, we insert $w$ to the sentence at index $i$, and continue guessing the $i + 1$ word based on the modified sentence.\footnote{We note that this process bears some similarity to Gibbs sampling from BERT conditioned LM.} 
We exclude a closed set of a few dozens of words (mostly function words) and keep them unchanged in all $k$ variations of a sentence. 
We further maintain structural correctness by maintaining the POS\footnote{We maintain the same POS so that the dataset will be valid for other tasks that require structure-preserving variants. However, In practice, we did not observe major differences when repeating the experiments reported here without the POS-preserving constraint when generating the data.}, and encourage semantic diversity by the auto-regressive replacement process. 
In Table \ref{tbl:parallel_examples} in the Appendix we show some additional generated groups.
The sets in Figure~\ref{fig:equivalent-sentences} were generated using this method.

\subsection{Word Representation}
We sample $N=150,000$ random sentences and use the our method to generate $900,000$ equivalent sets $E$ of structurally equivalent sentences. Then, we encode the sentences and randomly collect $1,500,000$ contextualized vector representations of words from these sets, resulting in 1,500,000 training pairs and 200,000 evaluation pairs for the training process of $f$. We experiment with both ELMo and BERT language models. In average, we sample 11 word-pairs from each group of equivalent sentences.
For ELMo, we represent each word in context as a concatenation of the last two ELMo layers (excluding the word embedding layer, which is not contextualized and therefore irrelevant for structure), resulting in representations of dimension 2048. For BERT, we concatenate the mean of the words' representation\footnote{Since BERT uses word-piece tokenization, we take the first token to represent each word.} across all contextualized layers of BERT-Large, with the representation of layer 16, which was found by \citet{structural-probe} most indicative of syntax.

\subsection{Triplet Loss}
\label{sec:triplet-loss}
We learn the mapping function $f$ using triplet loss (Figure \ref{fig:triplet-loss}). 
 Given a group of equivalent sentences $E_i$, we randomly choose two sentences to be the anchor sentence $S^A$ and the positive sentence  $S^P$, and sample two different word indices $\{i_1, i_2\}$. Let $S^A[i_1]$ be the contextualized representation of the $i_1$th word in sentence $S^A$. The words $S^A[i_1]$ and $S^A[i_2]$ from the anchor sentence would form a representation of a pair of words, which should be close to the pair $S^P[i_1]$, $S^P[i_2]$ from the positive sentence. 
 
 We represent pairs as their differences after transformation, resulting in the anchor pair $V^A$ and positive pair $V^P$:
 

\begin{align}
    V^A = f(S^A[i_1]) - f(S^A[i_2]) \;\;\;\;\;\; S^A \in E_i \\
    V^P = f(S^P[i_1]) - f(S^P[i_2]) \;\;\;\;\;\; S^P \in E_i
\end{align}
where $f$ is the parameterized syntactic transformation we aim to learn. 
We also consider a negative pair:
\begin{equation}
    V^N = 
    f(S^N[j_1]) - f(S^N[j_2])  \;\;\;\;\;\; S^N \not\in E_i
\end{equation}
coming from sentence $S^N$ which is not in the equivalent set.

As $f$ has shared parameters for both words in the pair, it can be considered a part of a Siamese network, making our learning procedure an instance of a triplet Siamese network \cite{triplet-siamese}. We choose $f$ to be a simple model: a single linear layer that maps from dimensionality 2048 to 75. The dimensions of the transformation were chosen according to development set performance. 

We use triplet loss \citep{triplet-siamese} to move the representation of the anchor vector $V^A$ closer to the representation of the positive vector $V^P$ and farther apart from the representation of the negative vector $V^N$. Following \citet{triplet-softmax}, we calculate the softmax version of the triplet loss:

\begin{equation}
L^{triplet}(V^A, V^P, V^N) = \frac{e^{d(V^A, V^P)}}{e^{d(V^A, V^P)} + e^{d(V^A, V^N)}}
\end{equation}

where $d(x,y) = 1 - \frac{x^{\T} y}{\|x\| \|y\|}$ is the cosine-distance between the vectors $x$ and $y$. 
The triplet objective is optimized end-to-end using the Adam optimizer \citep{adam}. We train for 5 epochs with a mini-batch of size 500 \footnote{A large enough mini-batch is necessary to find challenging negative examples.}, and take the last model as the final syntactic extractor. During training, the gradient backpropagates through the pair vectors to the parameters $f$ of the Siamese model, to get representations of individual words that are similar for corresponding words in equivalent sentences. We note that we do not back-propagate the gradient to the contextualized vectors: we keep them intact, and only adjust the learned transformation.

\paragraph{Hard negative sampling}
We obtain the negative vectors $V^N$ using hard negative sampling. For each mini-batch $B$, we collect 500 \{V$_i^A$, V$_i^P$\} pairs, each pair taken from an equivalent set $E_i$. The negative instances V$_i^N$ are obtained by searching the batch for a vector that is closest to the anchor and comes from a different set: 
\begin{equation}
V_i^N = \argmin_{\substack{V_{j\neq i}^A\in B}} d(V_i^A,V_j^A).
\end{equation}
In addition, we enforce a symmetry between the anchor and positive vectors, by adding a pair (positive, anchor) for each pair (anchor, positive) in $B$.
That is, $V_i^N$ is the ``most misleading'' word-pair vector: it comes from a sentence that has a different structure than the structure of V$_i^A$ sentence, but is the closest to V$_i^A$ in the mini-batch. 

\section{Experiments and Analysis}
\label{sec:experiments}

We have trained the syntactic transformation $f$ in a way that should encourage it to retain the structural information encoded in contextualized vectors, but discard other information. We assess the representations the model acquired in an unsupervised manner, by evaluating the extent to which the local neighbors of each transformed contextualized vector $f(x)$ share known structural properties, such as grammatical function within the sentence. For the baseline, we expect the neighbors of each vector to share a mix of semantic and syntactic properties. For the transformed vectors, we expect the neighbors to share mainly syntactic properties. Finally, we demonstrate that in a few-shot setting, our representations outperform the original ELMO representation, indicating they are indeed distilled from syntax, and discard other information that is encoded in ELMO vectors but is irrelevant for the extraction of the structure of a sentence. \\[0.5em]
\textbf{Corpus }
For training the transformation $f$, we rely on 150,000 sentences from Wikipedia, tokenized and POS-tagged by spaCy \cite{honnibal2015improved,honnibal2017spacy}. The POS tags are used in the equivalent set generation to filter replacement words. Apart from POS tagging, we do not rely on any syntactic annotation during training. The evaluation sentences for the experiments mentioned below are sampled from a collection of 1,000,000 original and unmodified Wikipedia sentences (different from those used in the model training). 

\subsection{Qualitative Analysis}
\begin{table*}[ht]
\resizebox{\textwidth}{!}{%

\begin{tabular}{c|l}
\toprule
Type & Text \\ \midrule
Q1 & \emph{in this way of thinking, an impacting \textbf{projectile} goes into an ice-rich layer -- but no further.} \\
N & they generally have a pre-engraved rifling band to engage the rifled launch tube, spin-stabilizing the \textbf{projectile}, hence the term ``rifle''. \\
NT & to achieve a large explosive yield, a linear implosion \textbf{weapon} needs more material, about 13 kgs. \\ \midrule

Q2 & \emph{the mint's \textbf{director} at the time, nicolas peinado, was also an architect and made the initial plans.} \\
N & the \textbf{director} is angry at crazy loop and glares at him, even trying to get a woman to kick crazy loop out of the show (which goes unsuccessfully). \\
NT & jetley's \textbf{mother}, kaushaliya rani, was the daughter of high court advocate shivram jhingan. \\ \midrule
Q3 & \emph{their first project is software that \textbf{lets} players connect the company's controller to their device}. \\
N & you could try use norton safe web, which \textbf{lets} you enter a website and show whether there seems to be anything bad in it.  \\
NT & the city offers a route-finding website that \textbf{allows} users to map personalized bike routes. \\

\bottomrule
\end{tabular}

}
\caption{Text examples for a few query words (in the Q rows, in bold), and their closest neighbours before (N) and after (NT) the transformation.}
\label{tbl:txt_example}

\end{table*}

\paragraph{t-SNE Visualization}

\begin{figure}[t!]
\centering
\subfloat{
\includegraphics[width=0.9\columnwidth]{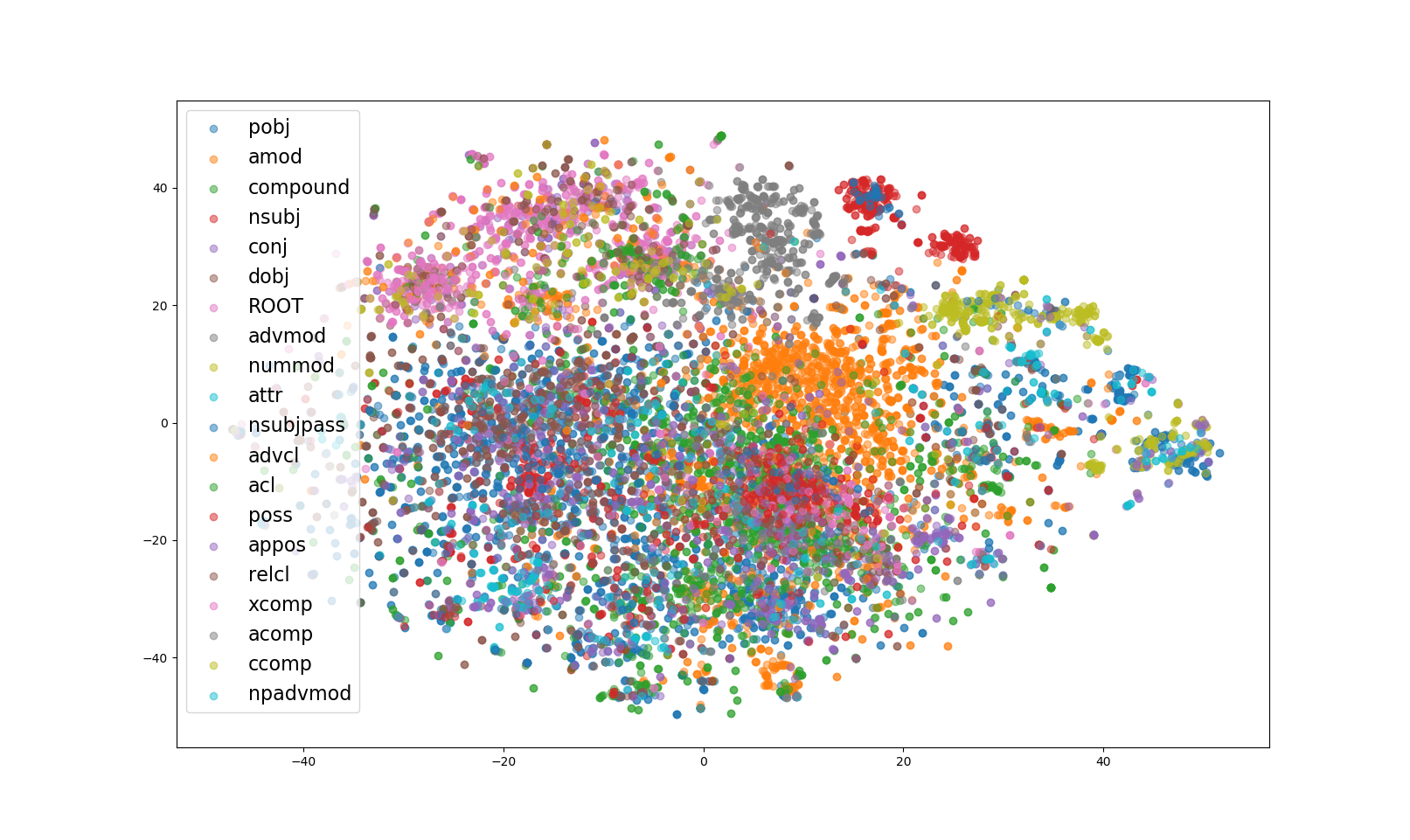}
}\\
\vspace{-8mm}
\subfloat{
    \includegraphics[width=0.9\columnwidth]{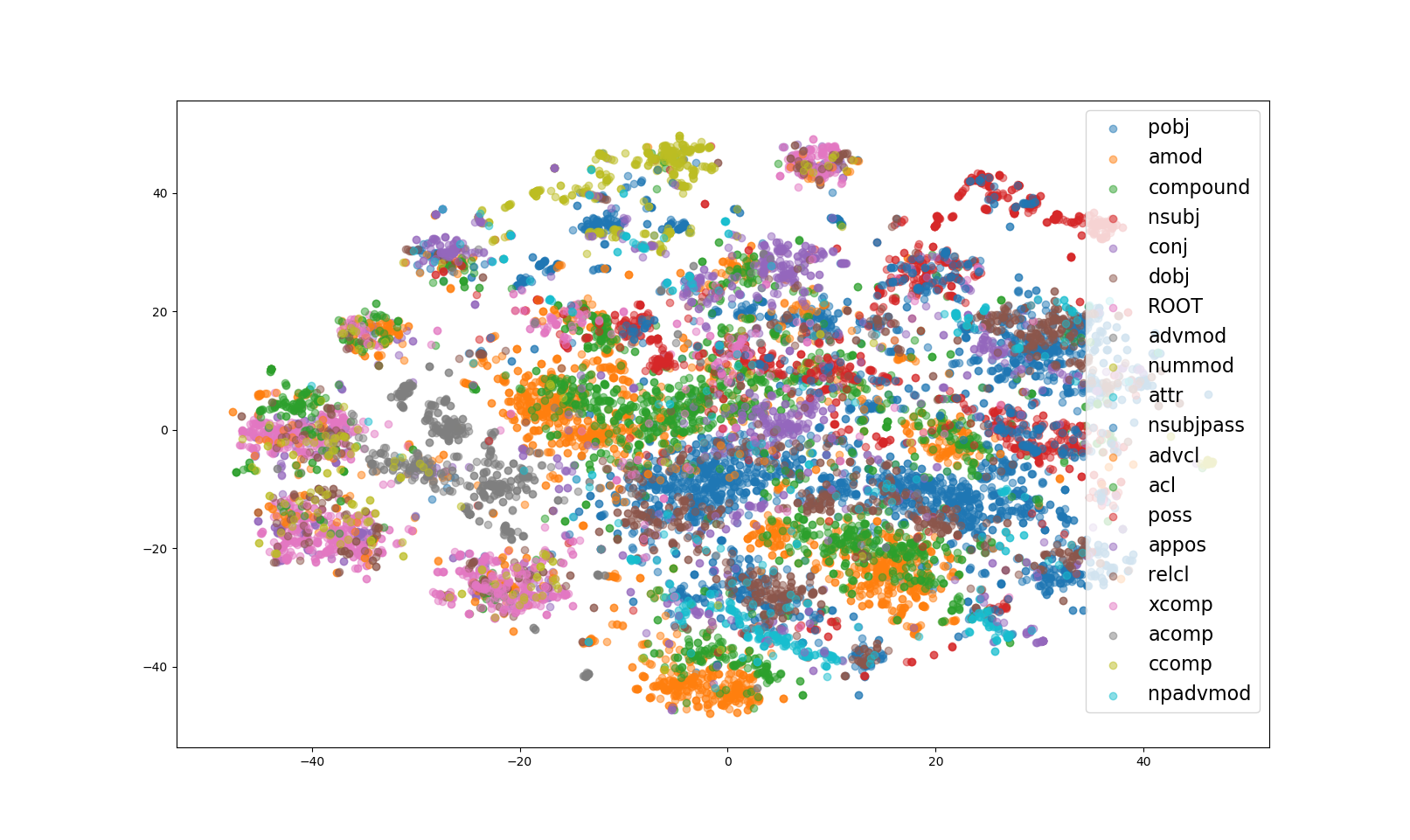}
}
\caption{t-SNE projection of ELMO states, colored by syntactic function, before (upper) and after (lower) the syntactic transformation.}
\label{fig:tsne}
\end{figure}
Figure \ref{fig:tsne} shows a 2-dimensional t-SNE projection \citep{t-sne} of 15,000 random content words. The left panel projects the original ELMo states, while the right panel is the syntactically transformed ones. The points are colored according to the dependency label (relation to parent) of the corresponding word, predicted by the parser.

In the original ELMo representation most states -- apart from those characterized by a specific part-of-speech, such as amod (adjectives, in orange) or nummod (numbers, in light green) -- do not fit well into a single cluster. In contrast, the syntactically transformed vectors are more neatly clustered, with some clusters, such as direct objects (brown) and prepositional-objects (blue), that are relatively separated after, but not before, the transformation. Interestingly, some functions that used to be a single group in ELMo (like the adjectives in orange, or the noun-compounds in green) are now split into several clusters, corresponding to their use in different sentence positions, separating for examples adjectives that are used in subject positions from those in object position or within prepositional phrases. Additionally, as noun compounds (``maple'' in ``maple syrup'') and adjectival modifiers (``tasty'' in ``tasty syrup'') are relatively structurally similar (they appear between determiners and nouns within noun phrases, and can move with the noun phrase to different positions), they are split and grouped together in the representation (the green and orange clouds). 

To quantify the difference, we run $K$-means clustering on the projected vectors, and calculate the average cluster purity score as the relative proportion of the most common dependency label in each cluster. The higher this value is, the more the division to clusters reflect division to grammatical functions (dependency labels). We run the clustering with different $K$ values: 10, 20, 40, 80. We find an increase in class purity following our transformation: from scores of 22.6\%, 26.8\%, 32.6\% and 36.4\% (respectively) for the original vectors, to scores of 24.3\%, 33.4\%, 42.1\% and 48.0\% (respectively) for the transformed vectors.



\paragraph{Examples} 
In Table \ref{tbl:txt_example} we present a few query words (Q) and their closest neighbours before (N) and after (NT) the transformation. Note the high structural similarity of the entire sentence, as well as the function of the word within it (Q1: last word of subject NP in a middle clause, Q2: possessed noun in sentence initial subject NP, Q3: head of relative clause of a direct object).

Additional examples (including cases in which the retrieved vector does not share the dependency edge with the query vector) are supplied in Appendix \S\ref{more-examples}.

\subsection{Quantitative Evaluation}
\label{sec:neighbors-test}

\begin{table*}[t!]
\centering
\resizebox{1\textwidth}{!}{%
\begin{tabular}{lccccccc} \hline
                   & Dep. edge & Head's dep. edge & Tree path  & Tree path & Tree path  & Depth & Lexical Match  \\
                   & & & (complete)       & (L=3)& (L=2) & (correlation) & \\
                   \hline
Baseline (all)       & 0.580     & 0.473            & 0.166                & 0.353           & 0.566           & 0.448    & 0.736           \\
Transformed (all)        & 0.699     & 0.603            & 0.253                & 0.523           & 0.735           & 0.561    & 0.284          \\ \hline

Transformed-untrained (all)   &   0.461   & 0.430 &   0.142   &   0.319   &   0.528   &   0.407   &   0.680 \\ \hline
Baseline (hard) & 0.509     & 0.460            & 0.160                & 0.347           & 0.564           & 0.430    & 0.776          \\
Transformed (hard)  & 0.671     & 0.591            & 0.260                & 0.534           & 0.751           & 0.576    & 0.274 \\   \hline

\end{tabular}
}
\caption{\label{tbl:closest_word_results} Closest-word queries, before and after the application of the syntactic transformation. ``Basline" refers to unmodified ELMo vectors,  ``Transformed" refers to ELMo vectors after the learned syntactic transformation  $f$, and ``Transformed-untrained" refers to ElMo vectors, after a transformation that was trained on a randomly-initialized ELMo. ``hard" denotes results on the subset of POS tags which are most structurally diverse.}

\end{table*}
We expect the transformed vectors to capture more structural and less lexical similarities than the source vectors. 
We expect each vectors' neighbors in space to share the structural function of the word over which the vector was collected, but not necessarily share its lexical meaning. We focus on the following structural properties: 
(1) Dependency-tree edge of a given word (dep-edge), that represents its function (subject, object etc.).
(2) The dependency edge of the word parent's (head's dep-edge) in the tree -- to represent higher level structure, such as a subject that resides within a relative clause, as in the word ``man" in the phrase ``the child that the man saw".
(3) Depth in the dependency tree (distance from the root of the sentence tree).
(4) Constituency-parse paths:  consider, for example, the sentence  ``They saw the moon with the telescope''. The word ``telescope" is a part of a noun-phrase ``the telescope", which resides inside a prepositional phrase ``with the telescope", which is part of the Verbal phrase ``saw with the telescope". The complete constituency path for this word is therefore ``NP-PP-VP". We calculate the complete tree path to the root (Tree-path-complete), as well as paths limited to lengths 2 and 3.
  

For this evaluation, we parse 400,000 random sentences taken from the 1-million-sentences Wikipedia sample, run ELMo and BERT to collect the contextualized representations of the sentences, and randomly choose 400,000 query word vectors (excluding function words). We then retrieve, for each query vector $x$, the value vector $y$ that is closest to $x$ in cosine-distance, and record the percentage of closest-vector pairs ($x, y$) that share each of the structural properties listed above. For the tree depth property, we calculate the Pearson correlation between the depths of the queries and the retrieved values. We use the Berkeley Neural Parser \citep{benepar-parser} for constituency parsing. 
We exclude function words from the evaluation.

\noindent
\paragraph{Easier and Harder cases}
The baseline models tend to retrieve words that are lexically similar. Since certain words tend to appear at above-chance probability in certain structural functions, this can make the baseline be ``right for the wrong reason'', as the success in the closest-word test reflects lexical similarity, rather than grammatical generalization. To control for this confounding, we sort the different POS tags according to the entropy of their dependency-labels distribution, and repeat the evaluation only for words belonging to those POS tags having the highest entropy (those are the most structurally variant, and tend to appear in different structural functions). The performance of the baselines (ELMo, BERT models) on those words drops significantly, while the performance of our model is only mildly influenced, indicating the superiority of the model in capturing structural rather than lexical information.








\paragraph{Results}

The results for ELMo are presented in Table \ref{tbl:closest_word_results}. For BERT, we witnessed similar, but somewhat lower, accuracy: for example, 68.1\% dependency-edge accuracy, 56.5\% head's dependency-edge accuracy, and 22.1\% complete constituency-path accuracy. The results for BERT are available in Appendix \S \ref{bert-results}, and for the reminder of the paper, we focus in ELMo. We observe significant improvement over the baseline for all tests. The correlation between the depth in tree of the query and the value words, for examples, rises from 44.8\% to 56.1\%, indicating that our model encourages the structural property of the depth of the word to be more saliently encoded in its representation compared with the baseline. The most notable relative improvement is recorded with regard to full constituency-path to the root: from 16.6\% before the structural transformation, to 25.3\% after it -- an improvement of 52\%. In addition to the increase in syntax-related properties, we observe a sharp drop --- from 73.6\% to 28.4\% --- in the proportion of query-value pairs that are lexically identical (lexical match, Table \ref{tbl:closest_word_results}). This indicates our transformation $f$ removes much of the lexical information, which is irrelevant for structure. To assess to what extent the improvements stems from the information encoded in ELMo, rather than being an artifact of the triplet-loss training, we also evaluate on a transformation $f$ that was trained on a randomly-initialized ELMo, a surprisingly strong baseline \citep{sent-vectors}. We find this model performs substantially worse than the baseline (Table \ref{tbl:closest_word_results}, ``Transformed-untrained (all)").  

\subsection{Minimal Supervision for Structure Distillation: Few-Shot Parsing}
\label{sec:parsing-test}


\begin{figure}[ht!]
\centering
\subfloat{
\centering
\includegraphics[width=0.8\columnwidth]{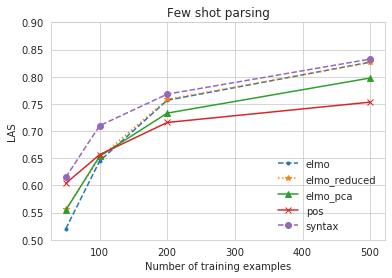}%
\label{fig:correlations-bert-biased}
} \\
\subfloat{
\centering
\includegraphics[width=0.8\columnwidth]{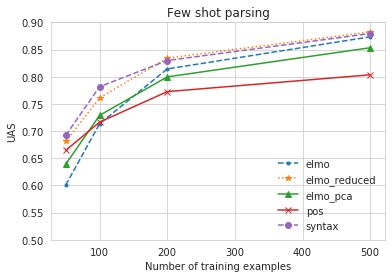}%
\label{fig:correlations-bert}
}
\caption{Results of the few-shots parsing setup.}
\label{fig:few-shot}
\end{figure}

The absolute nearest-neighbour accuracy values may appear to be relatively low: for example, only 67.6\% of the (query, value) pairs share the same dependency edge.

As the model acquires its representation without being exposed to human-mandated syntactic convention, some of the apparent discrepancies in nearest neighbours may be due to the fact the model acquires different kind of generalization, or learned a representation that emphasizes different kinds of similarities.
Still, we expect the resulting (75 dimensional) representations to contain distilled structure information that is mappable to human notions of syntax.
To test this, we compare dependency-parsers trained on our representation and on the source representation. 
If our representation indeed captures structural information, we expect it to excel on a low data setting. To this end, we test our hypothesis with few-shot dependency parsing setup, where we train a model to predict syntactic trees representation with only a few hundred labeled examples.

We use an off-the-shelf dependency parser model \citep{dozat2016deep} and swap the pre-trained Glove embeddings \citep{pennington2014glove} with ELMo contextualized embeddings \citep{elmo}. In order to have a fair comparison with our method, we use the concatenation of the two last layers of Elmo; we refer to this experiment as \textit{elmo}.
As our representation is much smaller than ELMo's (75 as opposed to 2048), a potential issue for a low data setting is the higher number of parameters to optimize in the later case, therefore a lower dimension may achieve better results.
We design two additional baselines to remedy this potential issue:
(1) Using PCA in order to reduce the representation dimensionality. We randomly chose 1M words from Wikipedia, calculated their representation with ELMo embeddings and performed PCA. This transformation is applied during training on top of ELMo representation while keeping the 75 first components. This experiment is referred to as \textit{elmo-pca}. This representation should perform well if the most salient information in the ELMo representations are structural. We exepct it to not be the case.
(2) Automatically learning a matrix that reduces the embedding dimension. This matrix is learned during training and can potentially extract the relevant structural information from the representations. We refer to this experiment as \textit{elmo-reduced}. 
Additionally, we also compare to a baseline where we use the gold-POS labels as the sole input to the model, by initializing an embedding matrix of the same size for each POS. We refer to this experiment as \textit{pos}.
Lastly, we examine the performance of our representation, where we apply our structural extraction method on top of ELMo representation. We refer to this experiment as \textit{syntax}. \\

We run the few-shot setup with multiple training size values: 50, 100, 200, 500. The results---for both labeled (LAS) and unlabeled (UAS) attachment scores---are presented in Figure \ref{fig:few-shot}, and the numerical results are available in the Appendix \S \ref{appendix-parsing}.
In the lower training size setting, we obtain the best performances compared to all baselines. The more training data is used, the gap between our representation and the baselines reduced, but the \emph{syntax} representation still outperforms \textit{elmo}. Using gold POS labels as inputs works relatively well with 50 training examples, but it quickly reaches a plato in performance and remains behind the other baselines. Reducing the dimensions with PCA (\textit{elmo-pca}) works considerably worse than ELMo, indicating PCA loses important information.  Reducing the dimensions with a learned matrix (\textit{elmo-reduced}) works substantially better than ELMo, and achieve the same UAS as our representation from 200 training sentences onward. However, our transformation was learned in an unsupervised fashion, without access to the syntactic trees. Finally, when considering the labeled attachment score, where the model is tasked at predicting not only the child-parent relation but also its label, our \textit{syntax} representation outperforms \textit{elmo-reduced}.



\section{Conclusion}

We propose an unsupervised method for the distillation of structural information from neural contextualized word representations. We used a process of sequential BERT-based substitution to create a large number of sentences which are structurally similar, but semantically different. By controlling for structure while changing lexical choice, we learn a metric under which pairs of words that come from structurally-similar sentences are close in space. We demonstrated that the representations acquired by this method share structural properties with their neighbors in space, and show that with a minimal supervision, those representations outperform ELMo in the task of few-shots parsing. The method is a first step towards a better disentanglement between various kinds of information that is represented in neural sequence models.

The method used to create the structurally equivalent sentences can be useful by its own as a data-augmentation technique. In future work, we aim to extend this method to allow for a more soft alignment between structurally-equivalent sentences.

\section*{Acknowledgments}
We would like to thank Gal Chechik for providing valuable feedback on early version of this work.
This project has received funding from the Europoean Research Council (ERC) under the Europoean Union's Horizon 2020 research and innovation programme, grant agreement No. 802774 (iEXTRACT).
Yanai Elazar is grateful to be partially supported by the PBC fellowship for outstanding PhD candidates in Data Science.


\bibliography{emnlp2020}

\begin{thebibliography}{49}
\expandafter\ifx\csname natexlab\endcsname\relax\def\natexlab#1{#1}\fi

\bibitem[{Adi et~al.(2016)Adi, Kermany, Belinkov, Lavi, and
  Goldberg}]{diagnostic1}
Yossi Adi, Einat Kermany, Yonatan Belinkov, Ofer Lavi, and Yoav Goldberg. 2016.
\newblock \href {http://arxiv.org/abs/1608.04207} {Fine-grained analysis of
  sentence embeddings using auxiliary prediction tasks}.
\newblock \emph{CoRR}, abs/1608.04207.

\bibitem[{Alemi et~al.(2016)Alemi, Fischer, Dillon, and
  Murphy}]{variational-IB}
Alexander Alemi, Ian Fischer, Joshua~V. Dillon, and Murphy Murphy. 2016.
\newblock Deep variational information bottleneck.
\newblock In \emph{Proceedings of the International Conference on Learning
  Representations (ICLR)}.

\bibitem[{Bengio et~al.(2013)Bengio, Courville, and
  Vincent}]{bengio-representation-learning}
Yoshua Bengio, Aaron~C. Courville, and Pascal Vincent. 2013.
\newblock \href {https://doi.org/10.1109/TPAMI.2013.50} {Representation
  learning: {A} review and new perspectives}.
\newblock \emph{{IEEE} Trans. Pattern Anal. Mach. Intell.}, 35(8):1798--1828.

\bibitem[{Conneau et~al.(2018)Conneau, Kruszewski, Lample, Barrault, and
  Baroni}]{sent-vectors}
Alexis Conneau, Germ{\'{a}}n Kruszewski, Guillaume Lample, Lo{\"{\i}}c
  Barrault, and Marco Baroni. 2018.
\newblock \href {https://doi.org/10.18653/v1/P18-1198} {What you can cram into
  a single {\textbackslash}{\textdollar}{\&}!{\#}* vector: Probing sentence
  embeddings for linguistic properties}.
\newblock In \emph{Proceedings of the 56th Annual Meeting of the Association
  for Computational Linguistics, {ACL} 2018, Melbourne, Australia, July 15-20,
  2018, Volume 1: Long Papers}, pages 2126--2136.

\bibitem[{Devlin et~al.(2019)Devlin, Chang, Lee, and Toutanova}]{bert}
Jacob Devlin, Ming{-}Wei Chang, Kenton Lee, and Kristina Toutanova. 2019.
\newblock \href {https://doi.org/10.18653/v1/n19-1423} {{BERT:} pre-training of
  deep bidirectional transformers for language understanding}.
\newblock In \emph{Proceedings of the 2019 Conference of the North American
  Chapter of the Association for Computational Linguistics: Human Language
  Technologies, {NAACL-HLT} 2019, Minneapolis, MN, USA, June 2-7, 2019, Volume
  1 (Long and Short Papers)}, pages 4171--4186. Association for Computational
  Linguistics.

\bibitem[{Dozat and Manning(2016)}]{dozat2016deep}
Timothy Dozat and Christopher~D Manning. 2016.
\newblock Deep biaffine attention for neural dependency parsing.
\newblock \emph{arXiv preprint arXiv:1611.01734}.

\bibitem[{Elazar et~al.(2020)Elazar, Ravfogel, Jacovi, and
  Goldberg}]{amnesic-probing:2020}
Yanai Elazar, Shauli Ravfogel, Alon Jacovi, and Yoav Goldberg. 2020.
\newblock \href {http://arxiv.org/abs/arXiv:2006.00995} {When bert forgets how
  to pos: Amnesic probing of linguistic properties and mlm predictions}.

\bibitem[{Ficler and Goldberg(2017)}]{ficler2017controlling}
Jessica Ficler and Yoav Goldberg. 2017.
\newblock Controlling linguistic style aspects in neural language generation.
\newblock \emph{arXiv preprint arXiv:1707.02633}.

\bibitem[{Fu et~al.(2018)Fu, Tan, Peng, Zhao, and Yan}]{fu2018style}
Zhenxin Fu, Xiaoye Tan, Nanyun Peng, Dongyan Zhao, and Rui Yan. 2018.
\newblock Style transfer in text: Exploration and evaluation.
\newblock In \emph{Thirty-Second AAAI Conference on Artificial Intelligence}.

\bibitem[{Gatys(2017)}]{image-style-transfer}
Leon~A. Gatys. 2017.
\newblock \href {http://d-nb.info/1151843520} {\emph{Texture synthesis and
  style transfer using perceptual image representations from convolutional
  neural networks}}.
\newblock Ph.D. thesis, University of T{\"{u}}bingen, Germany.

\bibitem[{Goldberg(2019)}]{goldberg2019}
Yoav Goldberg. 2019.
\newblock \href {http://arxiv.org/abs/1901.05287} {Assessing {BERT}'s syntactic
  abilities}.
\newblock \emph{CoRR}, abs/1901.05287.

\bibitem[{Gulordava et~al.(2018)Gulordava, Bojanowski, Grave, Linzen, and
  Baroni}]{gulordava2018LMagreement}
Kristina Gulordava, Piotr Bojanowski, Edouard Grave, Tal Linzen, and Marco
  Baroni. 2018.
\newblock \href {https://aclanthology.info/papers/N18-1108/n18-1108} {Colorless
  green recurrent networks dream hierarchically}.
\newblock In \emph{Proceedings of the Conference of the North American Chapter
  of the Association for Computational Linguistics: Human Language
  Technologies, {NAACL-HLT}}, pages 1195--1205.

\bibitem[{Hadad et~al.(2018)Hadad, Wolf, and Shahar}]{hadad-disentanglement}
Naama Hadad, Lior Wolf, and Moni Shahar. 2018.
\newblock \href {https://doi.org/10.1109/CVPR.2018.00087} {A two-step
  disentanglement method}.
\newblock In \emph{{IEEE} Conference on Computer Vision and Pattern
  Recognition, (CVPR)}, pages 772--780.

\bibitem[{Hewitt and Liang(2019)}]{Hewitt-control}
John Hewitt and Percy Liang. 2019.
\newblock \href {https://doi.org/10.18653/v1/D19-1275} {Designing and
  interpreting probes with control tasks}.
\newblock In \emph{Proceedings of the 2019 Conference on Empirical Methods in
  Natural Language Processing and the 9th International Joint Conference on
  Natural Language Processing, {EMNLP-IJCNLP} 2019, Hong Kong, China, November
  3-7, 2019}, pages 2733--2743. Association for Computational Linguistics.

\bibitem[{Hewitt and Manning(2019)}]{structural-probe}
John Hewitt and Christopher~D. Manning. 2019.
\newblock \href {https://aclweb.org/anthology/papers/N/N19/N19-1419/} {A
  structural probe for finding syntax in word representations}.
\newblock In \emph{Proceedings of the Conference of the North American Chapter
  of the Association for Computational Linguistics: Human Language
  Technologies, {NAACL-HLT}}, pages 4129--4138.

\bibitem[{Hoffer and Ailon(2015)}]{triplet-softmax}
Elad Hoffer and Nir Ailon. 2015.
\newblock \href {https://doi.org/10.1007/978-3-319-24261-3\_7} {Deep metric
  learning using triplet network}.
\newblock In \emph{Similarity-Based Pattern Recognition - Third International
  Workshop, {SIMBAD}}, pages 84--92.

\bibitem[{Honnibal and Johnson(2015)}]{honnibal2015improved}
Matthew Honnibal and Mark Johnson. 2015.
\newblock An improved non-monotonic transition system for dependency parsing.
\newblock In \emph{Proceedings of the 2015 conference on empirical methods in
  natural language processing}, pages 1373--1378.

\bibitem[{Honnibal and Montani(2017)}]{honnibal2017spacy}
Matthew Honnibal and Ines Montani. 2017.
\newblock spacy 2: Natural language understanding with bloom embeddings,
  convolutional neural networks and incremental parsing.
\newblock \emph{To appear}, 7(1).

\bibitem[{Hu et~al.(2017)Hu, Yang, Liang, Salakhutdinov, and
  Xing}]{HuControlledGeneration}
Zhiting Hu, Zichao Yang, Xiaodan Liang, Ruslan Salakhutdinov, and Eric~P. Xing.
  2017.
\newblock \href {http://proceedings.mlr.press/v70/hu17e.html} {Toward
  controlled generation of text}.
\newblock In \emph{Proceedings of the 34th International Conference on Machine
  Learning, {ICML} 2017, Sydney, NSW, Australia, 6-11 August 2017}, pages
  1587--1596.

\bibitem[{Hupkes et~al.(2018)Hupkes, Veldhoen, and Zuidema}]{diagnostic2}
Dieuwke Hupkes, Sara Veldhoen, and Willem~H. Zuidema. 2018.
\newblock \href {https://doi.org/10.1613/jair.1.11196} {Visualisation and
  'diagnostic classifiers' reveal how recurrent and recursive neural networks
  process hierarchical structure}.
\newblock \emph{J. Artif. Intell. Res.}, 61:907--926.

\bibitem[{Joshi et~al.(2019)Joshi, Levy, Weld, and Zettlemoyer}]{joshi2019bert}
Mandar Joshi, Omer Levy, Daniel~S Weld, and Luke Zettlemoyer. 2019.
\newblock {BERT} for coreference resolution: Baselines and analysis.
\newblock \emph{arXiv preprint arXiv:1908.09091}.

\bibitem[{Kingma and Ba(2015)}]{adam}
Diederik~P. Kingma and Jimmy Ba. 2015.
\newblock \href {http://arxiv.org/abs/1412.6980} {Adam: {A} method for
  stochastic optimization}.
\newblock In \emph{International Conference on Learning Representations,
  {ICLR}}.

\bibitem[{Kitaev and Klein(2018)}]{benepar-parser}
Nikita Kitaev and Dan Klein. 2018.
\newblock Constituency parsing with a self-attentive encoder.
\newblock In \emph{Proceedings of the Annual Meeting of the Association for
  Computational Linguistics (ACL)}.

\bibitem[{Lample et~al.(2018)Lample, Subramanian, Smith, Denoyer, Ranzato, and
  Boureau}]{lample2018multiple}
Guillaume Lample, Sandeep Subramanian, Eric Smith, Ludovic Denoyer,
  Marc'Aurelio Ranzato, and Y-Lan Boureau. 2018.
\newblock Multiple-attribute text rewriting.
\newblock In \emph{International Conference on Learning Representations}.

\bibitem[{Levy and Goldberg(2014)}]{levy2014linguistic}
Omer Levy and Yoav Goldberg. 2014.
\newblock Linguistic regularities in sparse and explicit word representations.
\newblock In \emph{Proceedings of the eighteenth conference on computational
  natural language learning}, pages 171--180.

\bibitem[{Li and Eisner(2019)}]{Eisner-IB}
Xiang~Lisa Li and Jason Eisner. 2019.
\newblock Specializing word embeddings (for parsing) by information bottleneck.
\newblock In \emph{Proceedings of the Conference on Empirical Methods in
  Natural Language Processing (EMNLP)}.

\bibitem[{Lin et~al.(2019)Lin, Tan, and Frank}]{lin2019open}
Yongjie Lin, Yi~Chern Tan, and Robert Frank. 2019.
\newblock Open sesame: Getting inside bert’s linguistic knowledge.
\newblock In \emph{Proceedings of the 2019 ACL Workshop BlackboxNLP: Analyzing
  and Interpreting Neural Networks for NLP}, pages 241--253.

\bibitem[{Liu et~al.(2019{\natexlab{a}})Liu, Gardner, Belinkov, Peters, and
  Smith}]{liu2019linguistic}
Nelson~F Liu, Matt Gardner, Yonatan Belinkov, Matthew Peters, and Noah~A Smith.
  2019{\natexlab{a}}.
\newblock Linguistic knowledge and transferability of contextual
  representations.
\newblock \emph{arXiv preprint arXiv:1903.08855}.

\bibitem[{Liu et~al.(2019{\natexlab{b}})Liu, Ott, Goyal, Du, Joshi, Chen, Levy,
  Lewis, Zettlemoyer, and Stoyanov}]{roberta}
Yinhan Liu, Myle Ott, Naman Goyal, Jingfei Du, Mandar Joshi, Danqi Chen, Omer
  Levy, Mike Lewis, Luke Zettlemoyer, and Veselin Stoyanov. 2019{\natexlab{b}}.
\newblock \href {http://arxiv.org/abs/1907.11692} {Ro{BERT}a: A robustly
  optimized {BERT} pretraining approach}.
\newblock \emph{CoRR}, abs/1907.11692.

\bibitem[{Maaten and Hinton(2008)}]{t-sne}
Laurens van~der Maaten and Geoffrey Hinton. 2008.
\newblock Visualizing data using t-{SNE}.
\newblock \emph{Journal of Machine Learning Research}, 9:2579--2605.

\bibitem[{Mathieu et~al.(2016)Mathieu, Zhao, Sprechmann, Ramesh, and
  LeCun}]{lecum-disentangling}
Micha{\"{e}}l Mathieu, Junbo~Jake Zhao, Pablo Sprechmann, Aditya Ramesh, and
  Yann LeCun. 2016.
\newblock \href
  {http://papers.nips.cc/paper/6051-disentangling-factors-of-variation-in-deep-representation-using-adversarial-training}
  {Disentangling factors of variation in deep representation using adversarial
  training}.
\newblock In \emph{Advances in Neural Information Processing Systems}, pages
  5041--5049.

\bibitem[{Maudslay et~al.(2020)Maudslay, Valvoda, Pimentel, Williams, and
  Cotterell}]{probe-parser}
Rowan~Hall Maudslay, Josef Valvoda, Tiago Pimentel, Adina Williams, and Ryan
  Cotterell. 2020.
\newblock \href {http://arxiv.org/abs/2005.01641} {A tale of a probe and a
  parser}.
\newblock \emph{CoRR}, abs/2005.01641.

\bibitem[{Mikolov et~al.(2013{\natexlab{a}})Mikolov, Sutskever, Chen, Corrado,
  and Dean}]{mikolov2013distributed}
Tomas Mikolov, Ilya Sutskever, Kai Chen, Greg~S Corrado, and Jeff Dean.
  2013{\natexlab{a}}.
\newblock Distributed representations of words and phrases and their
  compositionality.
\newblock In \emph{Advances in neural information processing systems}, pages
  3111--3119.

\bibitem[{Mikolov et~al.(2013{\natexlab{b}})Mikolov, Yih, and
  Zweig}]{mikolov2013linguistic}
Tomas Mikolov, Wen-tau Yih, and Geoffrey Zweig. 2013{\natexlab{b}}.
\newblock Linguistic regularities in continuous space word representations.
\newblock In \emph{Proceedings of the 2013 Conference of the North American
  Chapter of the Association for Computational Linguistics: Human Language
  Technologies}, pages 746--751.

\bibitem[{Narayanaswamy et~al.(2017)Narayanaswamy, Paige, van~de Meent,
  Desmaison, Goodman, Kohli, Wood, and Torr}]{digits-style}
Siddharth Narayanaswamy, Brooks Paige, Jan{-}Willem van~de Meent, Alban
  Desmaison, Noah~D. Goodman, Pushmeet Kohli, Frank~D. Wood, and Philip H.~S.
  Torr. 2017.
\newblock \href
  {http://papers.nips.cc/paper/7174-learning-disentangled-representations-with-semi-supervised-deep-generative-models}
  {Learning disentangled representations with semi-supervised deep generative
  models}.
\newblock In \emph{Advances in Neural Information Processing Systems}, pages
  5925--5935.

\bibitem[{Peng et~al.(2017)Peng, Yu, Sohn, Metaxas, and
  Chandraker}]{pose-invariant}
Xi~Peng, Xiang Yu, Kihyuk Sohn, Dimitris~N. Metaxas, and Manmohan Chandraker.
  2017.
\newblock \href {https://doi.org/10.1109/ICCV.2017.180} {Reconstruction-based
  disentanglement for pose-invariant face recognition}.
\newblock In \emph{{IEEE} International Conference on Computer Visionn (ICCV)},
  pages 1632--1641.

\bibitem[{Pennington et~al.(2014)Pennington, Socher, and
  Manning}]{pennington2014glove}
Jeffrey Pennington, Richard Socher, and Christopher Manning. 2014.
\newblock Glove: Global vectors for word representation.
\newblock In \emph{Proceedings of the 2014 conference on empirical methods in
  natural language processing (EMNLP)}, pages 1532--1543.

\bibitem[{Peters et~al.(2018)Peters, Neumann, Iyyer, Gardner, Clark, Lee, and
  Zettlemoyer}]{elmo}
Matthew~E. Peters, Mark Neumann, Mohit Iyyer, Matt Gardner, Christopher Clark,
  Kenton Lee, and Luke Zettlemoyer. 2018.
\newblock \href {https://doi.org/10.18653/v1/n18-1202} {Deep contextualized
  word representations}.
\newblock In \emph{Proceedings of the 2018 Conference of the North American
  Chapter of the Association for Computational Linguistics: Human Language
  Technologies, {NAACL-HLT} 2018, New Orleans, Louisiana, USA, June 1-6, 2018,
  Volume 1 (Long Papers)}, pages 2227--2237. Association for Computational
  Linguistics.

\bibitem[{Ravichander et~al.(2020)Ravichander, Belinkov, and
  Hovy}]{probing-probing}
Abhilasha Ravichander, Yonatan Belinkov, and Eduard~H. Hovy. 2020.
\newblock \href {http://arxiv.org/abs/2005.00719} {Probing the probing
  paradigm: Does probing accuracy entail task relevance?}
\newblock \emph{CoRR}, abs/2005.00719.

\bibitem[{Reif et~al.(2019)Reif, Yuan, Wattenberg, Viegas, Coenen, Pearce, and
  Kim}]{reif2019visualizing}
Emily Reif, Ann Yuan, Martin Wattenberg, Fernanda~B Viegas, Andy Coenen, Adam
  Pearce, and Been Kim. 2019.
\newblock Visualizing and measuring the geometry of bert.
\newblock In \emph{Advances in Neural Information Processing Systems}, pages
  8592--8600.

\bibitem[{Richardson et~al.(2019)Richardson, Hu, Moss, and
  Sabharwal}]{richardson2019probing}
Kyle Richardson, Hai Hu, Lawrence~S Moss, and Ashish Sabharwal. 2019.
\newblock Probing natural language inference models through semantic fragments.
\newblock \emph{arXiv preprint arXiv:1909.07521}.

\bibitem[{Schroff et~al.(2015)Schroff, Kalenichenko, and
  Philbin}]{triplet-siamese}
Florian Schroff, Dmitry Kalenichenko, and James Philbin. 2015.
\newblock \href {https://doi.org/10.1109/CVPR.2015.7298682} {{FaceNet}: {A}
  unified embedding for face recognition and clustering}.
\newblock In \emph{{IEEE} Conference on Computer Vision and Pattern Recognition
  (CVPR)}.

\bibitem[{Sohn et~al.(2015)Sohn, Lee, and Yan}]{sohn2015learning}
Kihyuk Sohn, Honglak Lee, and Xinchen Yan. 2015.
\newblock Learning structured output representation using deep conditional
  generative models.
\newblock In \emph{Advances in neural information processing systems}, pages
  3483--3491.

\bibitem[{Tenney et~al.(2019{\natexlab{a}})Tenney, Das, and
  Pavlick}]{nlp-bert-pipeline-tenney}
Ian Tenney, Dipanjan Das, and Ellie Pavlick. 2019{\natexlab{a}}.
\newblock \href {https://www.aclweb.org/anthology/P19-1452/} {{BERT}
  rediscovers the classical {NLP} pipeline}.
\newblock In \emph{Proceedings of the Conference of the Association for
  Computational Linguistics, {ACL}}, pages 4593--4601.

\bibitem[{Tenney et~al.(2019{\natexlab{b}})Tenney, Xia, Chen, Wang, Poliak,
  McCoy, Kim, Durme, Bowman, Das, and Pavlick}]{tenney2018what}
Ian Tenney, Patrick Xia, Berlin Chen, Alex Wang, Adam Poliak, R~Thomas McCoy,
  Najoung Kim, Benjamin~Van Durme, Sam Bowman, Dipanjan Das, and Ellie Pavlick.
  2019{\natexlab{b}}.
\newblock \href {https://openreview.net/forum?id=SJzSgnRcKX} {What do you learn
  from context? probing for sentence structure in contextualized word
  representations}.
\newblock In \emph{International Conference on Learning Representations}.

\bibitem[{Tishby et~al.(1999)Tishby, Pereira, and
  Bialek}]{tishby2000information}
Naftali Tishby, Fernando~C Pereira, and William Bialek. 1999.
\newblock The information bottleneck method.
\newblock In \emph{Proc. of the Allerton Allerton Conference on Communication,
  Control and Computing}.

\bibitem[{{van Schijndel} and Linzen(2018)}]{vanschijndel2018gardenpath}
Marten {van Schijndel} and Tal Linzen. 2018.
\newblock Modeling garden path effects without explicit hierarchical syntax.
\newblock In \emph{{Proceedings of the 40th Annual Conference of the Cognitive
  Science Society}}, pages 2600--2605. Cognitive Science Society.

\bibitem[{Wilcox et~al.(2018)Wilcox, Levy, Morita, and
  Futrell}]{wilcox2018fillergap}
Ethan Wilcox, Roger Levy, Takashi Morita, and Richard Futrell. 2018.
\newblock What do {RNN} language models learn about filler--gap dependencies?
\newblock In \emph{{Proceedings of the EMNLP Workshop BlackboxNLP: Analyzing
  and Interpreting Neural Networks for NLP}}, pages 211--221. Association for
  Computational Linguistics.

\bibitem[{Yang et~al.(2019)Yang, Xie, Lin, Li, Tan, Xiong, Li, and
  Lin}]{bert-questions}
Wei Yang, Yuqing Xie, Aileen Lin, Xingyu Li, Luchen Tan, Kun Xiong, Ming Li,
  and Jimmy Lin. 2019.
\newblock \href {https://aclweb.org/anthology/papers/N/N19/N19-4013/}
  {End-to-end open-domain question answering with {BERT}serini}.
\newblock In \emph{Proceedings of the Conference of the North American Chapter
  of the Association for Computational Linguistics {NAACL-HLT}}, pages 72--77.

\end{thebibliography}
\bibliographystyle{acl_natbib}

\clearpage
\appendix
\


\section{Additional Query-Value Examples}
\label{more-examples}
\begin{itemize}

\item
Q: \emph{as they did , the \textbf{probability} of an impact event temporarily climbed , peaking at 2 .}\\ 
N: \emph{however , the \textbf{probability} of flipping a head after having already flipped 20 heads in a row is simply } \\
NT: \emph{during the first year , the \textbf{scope} of red terror expanded significantly and the number of executions grew into the thousands .} 

\item

Q: \emph{the \textbf{celtics} honored his memory during the following season by retiring his number 35 .
}\\ 
N: \emph{the \textbf{beatles} performed the song at the 1969 let it be sessions .} \\
NT: \emph{the \textbf{warriors} dedicated their round five home match to fai 's memory .
} 

\item

Q: \emph{in the old zurich war , the swiss confederation plundered the monastery , whose \textbf{monks} had fled to zurich .
}\\ 
N: \emph{the hridaya sūtra and the `` five meditations '' are recited , after which \textbf{monks} will be served with the gruel and vegetables .} \\
NT: \emph{other commanders were killed and later rooplo kolhi was arrested near pag wool well , where his \textbf{troops} were fetching water.
}

\item

Q: \emph{the \textbf{main} cause of the punic wars was the conflict of interests between the existing carthaginian empire and the expanding roman republic .
}\\ 
N: \emph{the \textbf{main} issue was whether or not something had to be directly perceptible ( meaning intelligible to an ordinary human being ) for it to be a `` copy .} \\
NT: \emph{the \textbf{main} enemy of the game is a sadistic but intelligent arms-dealer known as the jackal , whose guns are fueling the violence in the country .} 

\item
Q: \emph{jones maintained lifelong links with his \textbf{native} county , where he had a home , bron menai , dwyran .
}\\ 
N: \emph{his association with the bbc ended in 1981 with a move back to his \textbf{native} county and itv company yorkshire television , replacing martin tyler as the regional station 's football commentator .} \\
NT: \emph{he leaves again for his \textbf{native} england , moving to a place near bath , where he works with a powerful local coven .
} 

\item
Q: \emph{silver iodate can be \textbf{obtained} by reacting silver nitrate ( agno3 ) with sodium iodate .
}\\ 
N: \emph{best mechanical strength is \textbf{obtained} if both sides of the disc are fused to the same type of glass tube and both tubes are under vacuum .} \\
NT: \emph{each of these options can be \textbf{obtained} with a master degree from the university along with the master of engineering degree .
} 

\item
Q: \emph{it \textbf{confirmed} that thomas medwin was a thoroughly learned man , if occasionally imprecise and careless 
}\\ 
N: \emph{it was \textbf{confirmed} that the truth about heather 's murder would be revealed which ultimately led to ben 's departure .} \\
NT: \emph{it \textbf{proclaimed} that the entire movement of plastic art of our time had been thrown into confusion by the discoveries above-mentioned .
} 

\item
Q: \emph{after the death of nadab and abihu , moses \textbf{dictated} what was to be done with their bodies .
}\\ 
N: \emph{most sources indicate that while no marriage took place between haile melekot and woizero ijigayehu , sahle selassie \textbf{ordered} his grandson legitimized .} \\
NT: \emph{vvkj pilots who flew the hurricane conversion \textbf{considered} it to be superior to the standard model .
} 

\item
Q: \emph{letters were delivered to sorters who \textbf{examined} the address and placed it in one of a number of `` pigeon holes '' .}\\ 
N: \emph{i \textbf{examined} and reported on the thread called transcendental meditation which appears on the page you linked to .} \\
NT: \emph{ronson visits purported psychopaths , as well as psychologists and psychiatrists who have \textbf{studied} them , and meets with robert d .
} 

\item
Q: \emph{slowboat to hades is a compilation \textbf{dvd} by gorillaz , released in october 2006 .}\\ 
N: \emph{the album was released in may 2003 as a single album with a bonus \textbf{dvd} .} \\
NT: \emph{master series is a compilation \textbf{album} by the british synthpop band visage released in 1997 .
} 

\item
Q: \emph{however , there are also many \textbf{theories} and conspiracies that describe the basis of the plot .}\\ 
N: \emph{the name tabasco is not definitively known with a number of \textbf{theories} debated among linguists} . \\
NT: \emph{it is likely that to this day there are some \textbf{harrisons} and harrises that are related .
}

\item
Q: \emph{nne , married first , to richard , eldest son of sir richard nagle , \textbf{secretary} of state for ireland , temp .}\\ 
N: \emph{in the early 1960s , profumo was the \textbf{secretary} of state for war in harold macmillan 's conservative government and was married to actress valerie hobson .} \\
NT: \emph{he was born in edinburgh , the son of william simpson , \textbf{minister} of the tron church , edinburgh , by his wife jean douglas balderston .
} 

\item
Q: \emph{battle of stoke field , the final \textbf{engagement} of the wars of the roses .}\\ 
N: \emph{among others , hogan announced the `` \textbf{engagement} '' of utah-born pitcher roy castleton .} \\
NT: \emph{song of susannah , the sixth \textbf{installment} in the dark tower series .
}

\item
Q: \emph{it vies for \textbf{control} with its host , causing physiological changes that will eventually cause the host 's internal organs to explode .}\\ 
N: \emph{hurtig and loewen developed rival factions within the party , and battled for \textbf{control} .} \\
NT: \emph{players take \textbf{control} of each of the four main characters at different times throughout the game , which enables multilateral perspective on the storyline .
} 

\item
Q: \emph{as such , radio tirana kept close to the official \textbf{policy} of the people 's republic of china , which was also both anti-west and anti-soviet whilst still being socialist in tone .}\\ 
N: \emph{this was in line with the \textbf{policy} outlined by constantine vii porphyrogenitus in de administrando imperio of fomenting strife between the rus ' and the pechenegs .
} \\
NT: \emph{april 2006 , the upr periodically examines the human rights \textbf{performance} of all 193 un member states .
} 

\item
Q: \emph{the engine was designed to \textbf{accept} either regular grade , 87 octane gasoline or premium grade , 91 octane gasoline .}\\ 
N: \emph{for example , an advanced html editing field could \textbf{accept} a pasted or inserted image and convert it to a data uri to hide the complexity of external resources from the user .
} \\
NT: \emph{it uses plug-ins ( html parsing technology ) to \textbf{collect} bibliographic information , videos and patents from webpages .}

\item
Q: \emph{one such decree was the notorious 1876 ems ukaz , which \textbf{banned} the kulishivka and imposed a russian orthography until 1905 ( called the yaryzhka , after the russian letter yery . ) .
}\\ 
N: \emph{fin 1612 , the shogun declared a decree that specifically \textbf{banned} the killing of cattle .
} \\
NT: \emph{tannis has \textbf{eliminated} the other time lords and set the doctor and the minister against each other .
}

\item
Q: \emph{a 25 degree list was \textbf{reduced} to 15 degrees ; men had abandoned ship prematurely - hence the pow .}\\ 
N: \emph{i suggest the article be \textbf{reduced} to something over half the size .} \\
NT: \emph{the old high school was \textbf{converted} into a middle school , until in 1971 the 5 .} 

\item
Q: \emph{the library catalog is maintained on a database that is \textbf{made} accessible to users through the internet.}\\ 
N: \emph{this screenshot is \textbf{made} for educational use and used for identification purposes in the article on nba on abc .} \\
NT: \emph{hpc is the main ingredient in cellugel which is \textbf{used} in book conservation .}

\item
Q: \emph{although he lost , \textbf{he} was evaluated highly by kazuyoshi ishii , and he was invited to seidokaikan .
}\\ 
N: \emph{\textbf{he} attended suny fredonia for one year and in 1976 received a b .} \\
NT: \emph{played primarily as a small forward , \textbf{he} showed some opportunist play and in his 18 games managed a creditable 12 goals .
}

\item
Q: \emph{for each \textbf{round} won , you gain one point towards winning the match .}\\ 
N: \emph{in the fourth \textbf{round} , federer beat tommy robredo and equalled jimmy connors ' record of 27 consecutive grand slam quarterfinals .} \\
NT: \emph{at the beginning of each \textbf{mission} , as well as the end of the last mission , a cutscene is played that helps develop the story .
}



\section{BERT Closest-Word Results}
\label{bert-results}

In Table \ref{tbl:closest_word_results_bert}, we present the full quantitative results when using BERT as the encoder. "Baseline" refers to unmodified vectors derived from BERT, and "Transformed" refers to the vectors after the learned syntactic transformation $f$. "hard" refers to evaluation on the subset of POS tags which are most structurally diverse.



\begin{table*}[ht!]
\resizebox{1.95\columnwidth}{!}{%
\begin{tabular}{lccccccc} \hline
                   & Dep. edge & Head's dep. edge & Tree path  & Tree path & Tree path  & Depth & Lexical Match  \\
                   & & & (complete)       & (L=3)& (L=2) & (correlation) & \\
                   \hline
Baseline (all)       & 0.549     & 0.432            & 0.146                & 0.310           & 0.522           & 0.436    & 0.829           \\
Transformed (all)        & 0.681     & 0.565            & 0.221                & 0.471           & 0.697           & 0.597    & 0.319          \\ \hline
Baseline (hard) & 0.478     & 0.429            & 0.143                & 0.310           & 0.521           & 0.428    & 0.820          \\
Transformed (hard)  & 0.652     & 0.565            & 0.225                & 0.482           & 0.714           & 0.601    & 0.300 \\   \hline

\end{tabular}
}

\caption{\label{tbl:closest_word_results_bert} Full quantitative results when using BERT as the encoder. "Baseline" refers to unmodified vectors derived from BERT, and "Transformed" refers to the vectors after the learned syntactic transformation $f$. "hard" refers to evaluation on the subset of POS tags which are most structurally diverse.}
\end{table*}

\section{Complete Parsing Results}
\label{appendix-parsing}
Below are the LAS and UAS scores for the experiments described in \S\ref{fig:few-shot}.

\label{parsing-numerical-results}

\begin{table}[ht]


\centering
\resizebox{\columnwidth}{!}{%
\begin{tabular}{lrrrr}
\hline
\multirow{-1}{*}{Model}                 & \multicolumn{4}{c}{Number of sentences}                                               \\
                 & 50 & 100 & 200 & 500  \\ \hline
POS              & 0.60 & 0.65 & 0.71 & 0.75 \\
ELMO             & 0.52 & 0.64 & 0.75 & 0.82 \\
ELMO-reduced     & 0.55 & 0.65 & 0.75 & 0.82 \\
ELMO-PCA         & 0.55 & 0.65 & 0.73 & 0.79 \\
ELMO-syntax (ours) & 0.61 & 0.70 & 0.76 & 0.83 \\
\end{tabular}
}
\label{tbl:parsing-las}
\caption{Labeled parsing scores (LAS)}


\end{table}

\begin{table}[ht]


\centering
\resizebox{\columnwidth}{!}{%
\begin{tabular}{lrrrr}
\hline
\multirow{-1}{*}{Model}                 & \multicolumn{4}{c}{Number of sentences}                                               \\
                    & 50   & 100 & 200  & 500  \\ \hline
POS                 & 0.66 & 0.71 & 0.77 & 0.80 \\
ELMO                & 0.60 & 0.71 & 0.81 & 0.87 \\
ELMO-reduced        & 0.68 & 0.76 & 0.83 & 0.88 \\
ELMO-PCA            & 0.63 & 0.72 & 0.79 & 0.85 \\
ELMO-syntax (ours) & 0.69 & 0.78 & 0.82 & 0.87
\end{tabular}
}

\label{tbl:parsing-uas}
\caption{Unlabeled parsing scores (UAS)}


\end{table}


\section{Examples of Equivalent Sentences}

In Table \ref{tbl:parallel_examples} we present randomly selected examples of groups of structurally-similar sentences (\S \ref{sec:sentences-generation}).

\label{equivalent-sentences-examples}

\begin{table*}[ht]
\centering
\resizebox{1.9\columnwidth}{!}{%

\begin{tabular}{ll}

\toprule
\#Version & Sentence \\
\midrule
Original & the structure is privately owned by the lake-hanford family of aurora , indiana and is not open to the public .\\ 
1 & the preserve is generally enjoyed by the ecological department of warren , california and is not free to the staff .\\ 
2 & the park is presently covered by the lake-hanford west of shrewsbury , italy and is not broken to the landscape .\\ 
3 & the festival is wholly offered by the west club of liberty , arkansas and is not central to the tradition .\\ 
4 & the pool is mostly administered by the shell town of greenville , maryland and is not navigable to the water .\\ 
5 & the house is geographically managed by the lake-hanford foundation of ferguson , fl and is not open to the sun .\\ 
\midrule
Original & on november 18th , 2011 , söllner released the studio album mei zuastand which features re-recorded songs from his entire career .\\ 
1 & on thursday 9th , 1975 , wolf dedicated the label das en imprint which comprises mixed albums from his golden series .\\ 
2 & on year 13th , 1985 , hoffmann wrote the vinyl mix von deutschland which plays imagined samples from his bible canon .\\ 
3 & on circa christmas , 2000 , press signed the lp debut re work which involves created phrases from his bible quote .\\ 
4 & on january 15th , 1995 , söllner wrote the camera y se theory which mixes cast phrases from his experimental archive .\\ 
5 & on oct 13th , 1983 , hansen organised the compilation concert ha radio which gives launched clips from his small film .\\ 
\midrule
Original & uhm ; we 're not proposing to give rollbackers the reviewer right .\\ 
1 & ah ; we 're not calling to quote comics the way hello .\\ 
2 & hi ; we 're not preparing to hear hits the dirt lady .\\ 
3 & shi ; we 're not asking to put rollbackers the board die .\\ 
4 & ar ; we 're not expecting to face rollbackers the place fell .\\ 
5 & whoa ; we 're not getting to detroit wants the boat paid .\\ 
\midrule
Original & coniston water is an example of a ribbon lake formed by glaciation .\\ 
1 & floating town is an artwork of a concrete area contaminated by mud .\\ 
2 & vista florida is an isle of a seaside lagoon fed by watershed .\\ 
3 & pit process is an occurrence of a hollow underground caused by settlement .\\ 
4 & union pass is an explanation of a highland section developed by anderson .\\ 
5 & ball phase is an exploration of a basalt basalt influenced by creep .\\ 
\midrule
Original & the highest lookout point , at above sea level , is trimble mountain , off brewer road .\\ 
1 & the greatest steep elevation , at above east cliff , is green rock , off little neck .\\ 
2 & the greatest lake club , at above east summit , is swiss cut , off northern pike .\\ 
3 & the biggest missing asset , at above single count , is local motel , off washington plaza .\\ 
4 & the smallest public surfing , at above virgin point , is grant lagoon , off white strait .\\ 
5 & the southwest east boundary , at above water flow , is trim hollow , off east town .\\ 
\midrule
Original & ample sdk is a lightweight javascript library intended to simplify cross-browser web application development .\\ 
1 & rapid editor is a popular editorial script suited to manage multi domain book edition .\\ 
2 & free id is a mandatory public implementation written to manage repository generic server environment .\\ 
3 & solar platform is a native developed stack written to ease regional complex sensing analysis .\\ 
4 & standard library is a complete python interface required to provide cellular mesh construction engine .\\ 
5 & flex module is a standardized foundry block applied to facilitate component development common work .\\ 
\midrule
Original & she wore a pale pink gown , silver crown and had pale pink wings .\\ 
1 & she boasted a large halt purple , fuzzy lip and had twin firm wrists .\\ 
2 & she spun a thin olive jelly , joined yarn and had large silver bubbles .\\ 
3 & she flared a high frequency yellow , reddish rose and had fried like moses .\\ 
4 & she exhibited a small frame overall , broad head and had oval eyed curves .\\ 
5 & she wrapped a silky ga yellow , moth hide and had homemade gold roses .\\ 
\midrule
Original & tegan is somewhat quiet and is rather scared , but kamryn reasures her everything will be ok .\\ 
1 & man is slightly pissed and is rather awkward , but kamryn protests her night will be ok .\\ 
2 & lao is real sad and is rather disappointed , but san figures her story will be ok .\\ 
3 & daughter is increasingly pregnant and is rather uncomfortable , but ni confirms her birth will be ok .\\ 
4 & mai is strangely warm and is rather short , but papa wishes her day will be ok .\\ 
5 & mare is slowly back and is rather upset , but pa asserts her sister will be ok .\\ 
\midrule
Original & shapley participated in the `` great debate '' with heber d .\\ 
1 & morris put in the `` heroic speech '' with heber energy .\\ 
2 & hall met in the `` ninth season '' with walton moore .\\ 
3 & patel helped in the `` double coup '' with ibn salem .\\ 
4 & chu sent in the `` universal text '' with u z .\\ 
5 & smith exhibited in the `` red year '' with william james .\\ 
\midrule
Original & the added english voice-over narration by the vampire ancestor removes any ambiguity .\\ 
1 & the untitled thai adventure script by the light corps includes any future .\\ 
2 & the improved industrial hole tool by the freeman workshop touches any resistance .\\ 
3 & the arched robotic interference use by the computer computer checks any message .\\ 
4 & the fixed regular speech described by the german army encompasses any type .\\ 
5 & the combined complete phone acquisition by the surround computer marks any microphone .\\ 

\bottomrule
\end{tabular}
}
\caption{Randomly selected examples of groups of structurally-similar sentences (\S \ref{sec:sentences-generation})}
\label{tbl:parallel_examples}

\end{table*}

\end{itemize}

\end{document}